%% file: root.tex
\documentclass[letterpaper, 10 pt, journal, twoside]{IEEEtran}

\IEEEoverridecommandlockouts                              

\usepackage{graphics} 
\usepackage{epsfig} 
\usepackage{times} 
\usepackage{amsmath} 
\usepackage{amssymb}  
\usepackage{hyperref} 
\usepackage{xcolor} 
\usepackage{color} 
\usepackage{float}
\usepackage{multirow}
\usepackage{siunitx}
\usepackage{textcomp}
\usepackage{float}
\usepackage{multirow}
\usepackage{booktabs}
\usepackage{wrapfig}
\usepackage{tabularx}
\usepackage{bm}
\usepackage{tikz}
\usepackage{tikz-3dplot}
\usetikzlibrary{calc,math, positioning, intersections, decorations, external, patterns, arrows,perspective, shapes.geometric}
\usepackage{pgfmath}
\usepackage{pgfplots}
\usepackage{xifthen}
\usepackage{listofitems}
\usepackage{array}

\pgfplotsset{compat=1.16}
\usepackage[skip=0pt,font=footnotesize]{caption}
\usepackage{subcaption}
\newcommand{\rebuttal}[1]{{\color{black} #1}}

\newcommand\norm[1]{\left\lVert#1\right\rVert}

\definecolor{matlab1}{rgb}{0.00000,0.44700,0.74100}%
\definecolor{matlab2}{rgb}{0.85000,0.32500,0.09800}%
\definecolor{matlab3}{rgb}{0.92900,0.69400,0.12500}%
\definecolor{matlab4}{rgb}{0.49400,0.18400,0.55600}%
\definecolor{matlab5}{rgb}{0.4660, 0.6740, 0.1880}%
\definecolor{matlab6}{rgb}{0.3010, 0.7450, 0.9330}%

\newcommand{\rotateRPY}[3]
{   \pgfmathsetmacro{\rollangle}{#1}
    \pgfmathsetmacro{\pitchangle}{#2}
    \pgfmathsetmacro{\yawangle}{#3}

    \pgfmathsetmacro{\newxx}{cos(\yawangle)*cos(\pitchangle)}
    \pgfmathsetmacro{\newxy}{sin(\yawangle)*cos(\pitchangle)}
    \pgfmathsetmacro{\newxz}{-sin(\pitchangle)}
    \path (\newxx,\newxy,\newxz);
    \pgfgetlastxy{\nxx}{\nxy};

    \pgfmathsetmacro{\newyx}{cos(\yawangle)*sin(\pitchangle)*sin(\rollangle)-sin(\yawangle)*cos(\rollangle)}
    \pgfmathsetmacro{\newyy}{sin(\yawangle)*sin(\pitchangle)*sin(\rollangle)+ cos(\yawangle)*cos(\rollangle)}
    \pgfmathsetmacro{\newyz}{cos(\pitchangle)*sin(\rollangle)}
    \path (\newyx,\newyy,\newyz);
    \pgfgetlastxy{\nyx}{\nyy};

    \pgfmathsetmacro{\newzx}{cos(\yawangle)*sin(\pitchangle)*cos(\rollangle)+ sin(\yawangle)*sin(\rollangle)}
    \pgfmathsetmacro{\newzy}{sin(\yawangle)*sin(\pitchangle)*cos(\rollangle)-cos(\yawangle)*sin(\rollangle)}
    \pgfmathsetmacro{\newzz}{cos(\pitchangle)*cos(\rollangle)}
    \path (\newzx,\newzy,\newzz);
    \pgfgetlastxy{\nzx}{\nzy};
}

\newcolumntype{C}{>{\centering\arraybackslash}X}
\newcolumntype{P}[1]{>{\centering\arraybackslash}p{#1}}

\title{Drift-free Visual SLAM using Digital Twins}

\author{Roxane Merat$^{\star}$, Giovanni Cioffi$^{\star}$, Leonard Bauersfeld, and Davide Scaramuzza
\thanks{Manuscript received: August, 30, 2024; Revised November, 1, 2024; Accepted November, 25, 2024.}
\thanks{
This paper was recommended for publication by Editor Javier Civera upon evaluation of the Associate Editor and Reviewers' comments.}
\thanks{This work was supported by the European Union’s Horizon-Europe Research and Innovation Programme under grant agreement No. 101120732 (AUTOASSESS) and by the European Research Council (ERC) under grant agreement No. 864042 (AGILEFLIGHT).
}%
\thanks{$^{\star}$These authors contributed equally. The authors are with the Robotics and Perception Group, Department of Informatics, University of Zurich, and Department of Neuroinformatics, University of Zurich and ETH Zurich, Switzerland (\protect\url{http://rpg.ifi.uzh.ch}).}%
\thanks{Digital Object Identifier (DOI): see top of this page.}
}

\begin{document}

\maketitle

\begin{abstract}

Globally-consistent localization in urban environments is crucial for autonomous systems such as self-driving vehicles and drones, as well as assistive technologies for visually impaired people. 
Traditional Visual-Inertial Odometry (VIO) and Visual Simultaneous Localization and Mapping (VSLAM) methods, though adequate for local pose estimation, suffer from drift in the long term due to reliance on local sensor data. 
While GPS counteracts this drift, it is unavailable indoors and often unreliable in urban areas.
An alternative is to localize the camera to an existing 3D map using visual-feature matching. 
This can provide centimeter-level accurate localization but is limited by the visual similarities between the current view and the map.
This paper introduces a novel approach that achieves accurate and globally-consistent localization by aligning the sparse 3D point cloud generated by the VIO/VSLAM system to a digital twin using point-to-plane matching; no visual data association is needed. 
The proposed method provides a 6-DoF global measurement tightly integrated into the VIO/VSLAM system. 
Experiments run on a high-fidelity GPS simulator and real-world data collected from a drone demonstrate that our approach outperforms state-of-the-art VIO-GPS systems and offers superior robustness against viewpoint changes compared to the state-of-the-art Visual SLAM systems.

\end{abstract}

\begin{IEEEkeywords}
SLAM; Localization; Mapping
\end{IEEEkeywords}

\input{sections/supplementary_material}
\input{sections/introduction}
\input{sections/related_work}

\input{sections/methodology}
\input{sections/experiments}
\input{sections/discussion}
\input{sections/conclusion}

{\small
\bibliographystyle{IEEEtran}
\bibliography{bibliography}
}

\end{document}

%% file: sections/supplementary_material.tex
\section*{Supplementary Material}\label{sec:SupplementaryMaterial}

\textbf{Video}: \url{https://youtu.be/gmHnhWYfuW0}

\textbf{Code}:\url{https://github.com/uzh-rpg/rpg_svo_pro_with_digital_twins}

%% file: sections/introduction.tex
\section{Introduction}\label{sec:Introduction}



\IEEEPARstart{A}{ccurate} and globally-consistent pose estimation in complex environments is a critical requirement for autonomous systems~\cite{eu}, such as self-driving vehicles and drones~\cite{badue2021self}, as well as assistive technologies for visually-impaired people~\cite{gionfrida2024wearable}.
Camera and inertial measurement units (IMU) are commonly used to solve the pose estimation~\cite{cadena2016past,huang2019visual} problem thanks to their low cost, lightweight design, and complementary measurement capabilities—cameras providing rich but low-rate and environment-dependent data, and IMUs offering high-rate, low-dimensional, and environment-independent data.
\begin{figure}
    \centering
    \includegraphics[width=1.0\linewidth]{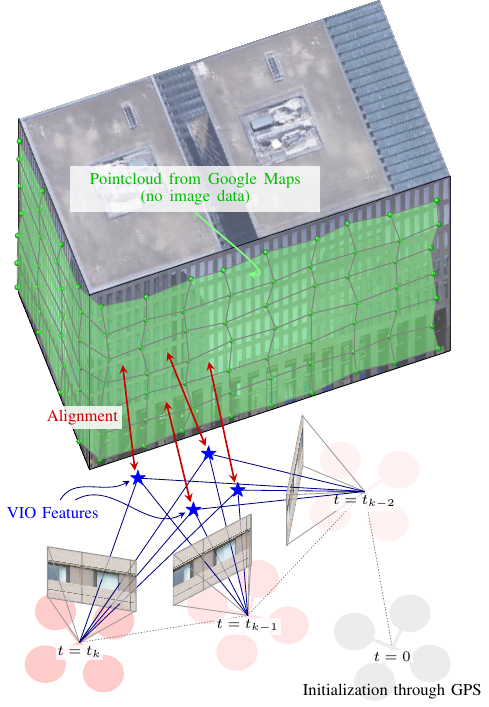}
    \caption{We propose an approach to achieve drift-free visual SLAM by aligning the local visual SLAM point cloud to a digital twin using point-to-plane matching. The resulting relative transformation between the point cloud and the digital twin provides a global measurement that is then tightly integrated into the SLAM system to obtain global consistency and reduce drift.}
    \label{fig:enter-label}
\end{figure}
Visual-Inertial Odometry (VIO) and Visual Simultaneous Localization and Mapping (VSLAM) algorithms fuse camera and IMU measurements and output the pose estimate in a local frame. The local frame is usually set equal to the identity pose at the initialization of the algorithm.
Due to the fact that camera and IMU provide local measurements, as well as noise and modeling simplifications, these algorithms accumulate drift over time.

To obtain pose estimates in a global fixed frame (e.g., East-North-Up (ENU) reference frame), and to reduce the drift, global measurements are included in the pose estimation process~\cite{cioffi2020tightly}.
The most commonly used global measurement is GPS (Global Positioning System). However, GPS measurements are unavailable indoors and often noisy and unreliable in urban canyons, where signal reflections and blockages are common. As a result, GPS measurements can sometimes be more uncertain than the pose estimates generated by the VIO/VSLAM system itself.

VSLAM systems compensate for the drift by localizing to a pre-built map using visual measurements.
The localization problem is solved by matching visual features between the current camera view and the map.
Standard techniques match feature descriptors~\cite{mur2015orb, campos2021orb, Oertel}.
More recent works, instead, replace feature descriptors with neural networks~\cite{sarlin2023orienternet, panek2023visual, sarlin2024snap, Ye}.
Regardless of the matching technique, all the visual-based localization methods rely on the similarity between the current camera view and the one in the reference map, limiting their effectiveness in many situations, such as localizing aerial images (e.g., taken by a drone) to a map recorded from a ground vehicle during day and night, or localizing a VIO/VSLAM system based on other spectra cameras (e.g., infrared cameras) to a map recorded with a standard camera.


Despite recent advancements in VIO-GPS algorithms and visual localization techniques, existing methods still struggle to fully address the challenge of achieving globally consistent, low-drift pose estimation in urban environments due to the inherent noise, sparsity, or viewpoint dependency of the global measurements on which they rely on. 

In response to these challenges, we propose to use geometric information (e.g., 3D point clouds, meshes) to localize to digital twins. Notably, we introduce a novel approach that localizes the local map (sparse 3D point cloud) generated by the VIO/VSLAM systems to the geometric digital twin. Our method uses the local VIO/VSLAM point cloud to perform point-to-plane matching against the mesh representation of the digital twin, providing a new global measurement that is tightly fused into the VIO/VSLAM system.

Our experiments demonstrate the superiority of the proposed method over the current state-of-the-art VIO-GPS approaches, both in a high-fidelity GPS simulator, which we have developed in the context of this project, and real-world tests conducted with a drone flying in a city. 
The results show that our approach not only reduces drift more effectively but also offers greater robustness to variations in viewpoint compared to traditional visual localization techniques based on feature descriptors matching. 
The main contributions of this work are:
\begin{itemize}
    \item A novel algorithm that localizes the local and sparse VSLAM 3D point cloud to a digital twin via point-to-plane matching, consequently being independent of visual matching. 
    \item Tightly fusing such localization measurement in a VSLAM system to achieve globally consistent, low-drift pose estimates in urban environments.
    \item We show the superiority of our approach against state-of-the-art VIO-GPS systems in simulation experiments using a novel high-fidelity GPS simulator and real-world data collected onboard a drone flying in a city.
    \item We show that the proposed local-global map localization is more robust than the state-of-the-art vision localization technique based on image feature descriptors matching.
\end{itemize}

%% file: sections/related_work.tex
\section{Related Work}\label{sec:Related_work}

Due to drift, long-term pose estimation in large-scale environments requires global measurements, i.e. measurements relative to a static frame.

GPS measurements are widely used in VIO/VSLAM systems to achieve low drift pose estimates in the long term.
Early efforts in this direction include the work of Qin et al.~\cite{qin2019general} and Mascaro et al.~\cite{MascaroGOMSF}, who loosely fuse GPS data with a visual-inertial odometry system. 
In subsequent work, Cioffi et al.~\cite{cioffi2020tightly} show that the accuracy of the pose estimates can be improved by tightly coupling visual, inertial, and GPS measurements.
Boche et al.~\cite{boche2022visualinertial} extend the tightly-coupled approach to make it robust to short-term GPS data dropouts.
All these methods rely on the availability of GPS data and consequently work well in open-sky scenarios but suffer in urban environments.
Different methods have been proposed to deal with urban environments. 
Cao et al.~\cite{cao2021gvins} propose a raw GPS pseudorange measurement model, which improves the system's resilience against corrupted GPS signals. Yin et al.~\cite{SkyGVINS} add filtering of these measurements based on an upward-facing camera, keeping only the pseudoranges from line-of-sight satellites. Lee et al.~\cite{Lee2020IntermittentGV} propose a method that is resilient to the loss of GPS signals by gradually adding GPS measurements when they become available again. 
Although these methods help to limit the error introduced by uncertain GPS measurements in dense areas, they suffer when GPS measurements are corrupted for long periods.

Alternative methods obtain global measurements by localizing to a map of the environment.
In VSLAM, the localization process is based on visual data association between the current camera view and the map.
Classical loop closure techniques rely on matching feature descriptors, e.g., the ORB descriptor~\cite{mur2015orb}.
Majdik et al.~\cite{CADZurich} have developed a method to match images captured by unmanned aerial vehicles (UAVs) with Google Street View images by back-projecting their textures onto CAD models of cities to infer global positions. 
In a follow-up work~\cite{Majdik_2}, they improved their approach against viewpoint differences. While this method can compensate for moderate viewpoint differences, UAVs can navigate at altitudes where the camera field of view does not overlap with the Google Street View images.
Bao et al.~\cite{Bao} propose a similar approach as~\cite{CADZurich, Majdik_2} where they first back-project the VIO point cloud onto a previous VSLAM-generated map, then use visual feature matching to verify these projections and derive a localization constraint.
Recently, many methods relying on deep learning have been proposed~\cite{sarlin2023orienternet, panek2023visual, sarlin2024snap}.
These methods usually train a neural network to predict the pose of the camera given an image. The training data usually consists of images contained in the map to which one is interested to localize to.
We refer the reader to the survey paper in~\cite{chen2023deep} where a broad overview of deep learning techniques for visual localization is provided.
Satellite images have also been used to localize the current camera view in a fixed global frame by Sadli et al.~\cite{SADLI2022255} where they propose a technique to match vehicle images with satellite images by aligning road features.
Visual localization techniques provide cm-accurate localization when visual data association can be accurately performed. However, they are very susceptible to viewpoint differences and environmental conditions (e.g., lighting).
For example, methods that rely on Google Street View images, which are taken from ground viewpoints, fail to localize images taken onboard UAVs flying high above the ground.

In this work, we are inspired by point cloud registration techniques, which are at the core of LiDAR-based SLAM systems.
Point cloud registration aims to find the transformation that aligns two point clouds.
Commonly, this problem is solved by applying the iterative closest point (ICP) algorithm~\cite{vizzo2023kiss}.
Indeed, we propose to localize the local point cloud generated by a VIO/VSLAM system to a mesh representation of a city digital twin.

%% file: sections/methodology.tex
\section{Methodology}\label{sec:methodology}
\begin{figure}
    \centering
    \includegraphics[width=1.0\linewidth]{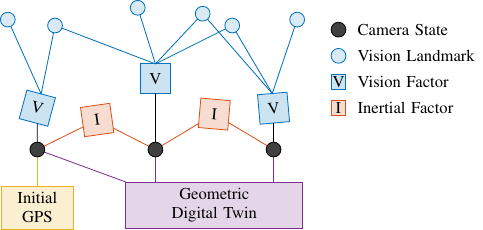}
    \vspace{0.5cm}
    \caption{Factor graph representation of the proposed visual SLAM system with visual, inertial, and city digital twin registration factors. The system relies on GPS data to initialize.}
    \label{fig:factor_graph}
\end{figure}

In this work, we propose a VSLAM system that achieves low-drift and global consistent pose estimates in urban environments by leveraging the scene geometry to localize to a digital twin.
The localization problem is formulated as a point cloud registration problem that can be efficiently solved by the point-to-plane ICP algorithm.
Specifically, we obtain a global measurement by registering the local sparse 3D point cloud generated by the VSLAM system to the city digital twin.
This global measurement is then tightly coupled with the current camera and IMU measurements in the VSLAM pose estimation process.
Additionally, we propose an algorithm that adaptively finds the weights that are used to fuse the global measurement in the VSLAM. 
This adaptive weighting solution allows us to account for uncertain solutions to the registration problem, which derive from degenerate cases and depend on the scene geometry.
We describe in this section our methodology in detail. 
A diagram of our methodology is in Fig.~\ref{fig:factor_graph}

\subsection{Notation}

\begin{figure}
    \centering
    \includegraphics[width=1.0\linewidth]{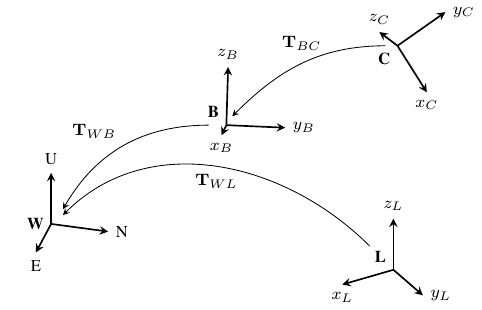}
    \caption{Reference frames used in this work.}
    \label{fig:reference_frames}
\end{figure}

The coordinates frames used in this work are illustrated in Fig.~\ref{fig:reference_frames}. 
We transform the GPS measurements, typically represented in the geodetic coordinate system (latitude, longitude, altitude), in Cartesian coordinates (x, y, z) expressed in the ENU frame, denoted by $W$. 
$L$ represents the VSLAM local frame. This frame is initialized with the identity at the first VSLAM pose estimate. 
$B$ is the body frame and corresponds to the IMU frame. 
The camera frame is denoted by $C$. 
We assume that the IMU-camera calibration is known.
To simplify the mathematical derivations in this section, we assume that the GPS and camera-IMU sensor platform are time synchronized and that the GPS antenna is located in the same place as the IMU. These quantities can be obtained in practice via offline calibration~\cite{furgale2013unified, cioffi2022continuous}. 
We use $\bm{p}_{WB}^{k}$ and $\bm{R}_{WB}^{k}$ to represent the position and orientation (in the form of a rotation matrix as an element of \textit{SO}(3)) of the frame $B$ with respect to $W$ at time $t_k$.
The corresponding homogeneous transformation matrix is written as $\bm{T}_{WB}^{k} = [\bm{R}_{WB}^{k} | \bm{p}_{WB}^{k}]$.
The same notation applies to the other reference frames.
We use the symbol $\hat{\cdot}$ to represent the estimated quantities, for example, $\hat{\bm{p}}_{WB}^{k}$ is the position estimated by the VSLAM at time $t_k$.

\subsection{Problem Formulation}

Our pose estimation problem follows the formulation of the sliding-window optimization-based VSLAM algorithm~\cite{huang2019visual}.
The camera poses, as well as the local sparse 3D point cloud, and the IMU biases, are obtained by minimizing the following cost function:
\begin{equation} \label{Eq:VIO_Formulation}
\begin{split}
    \min_{\mathcal{X}} \mathcal{J} = \min_{\mathcal{X}} & \sum_{k=0}^{K-1} \sum_{j\in\mathcal{J}_k} \left\| \bm{e}_v^{j,k} \right\|^2_{W_v^{j,k}} 
    + \sum_{k=0}^{K-1} \left\| \bm{e}_i^{k} \right\|^2_{W_i^k} \\
    &+ \sum_{k=0}^{K-1}  \left\| \bm{e}_{m}^{k} \right\|^2_{W_m^{k}}
    + \left\| \bm{e}_p \right\|^2  
\end{split}
\end{equation}
where $\mathcal{X}$ is the vector that contains all the optimization variables.
%

%
The term $\bm{e}_{v}^{j,k} = \bm{z}^{j,k} - h(\bm{l}_{j}^{l})$ represents the visual factor.
Given $\bm{l}_{j}^{l} \in \mathcal{J}_k$ a 3D point visible from the keyframe $k$, the visual factor is defined by reprojecting $\bm{l}_{j}^{l}$ to the image frame via the function $h(\cdot)$ and computing the error against the corresponding image point $\bm{z}^{j,k}$.
The term $\bm{e}_{i}$ represents the inertial factor.
To compute this factor, we use the IMU preintegration algorithm proposed by Forster et al.~\cite{forster2016manifold}.
The term $\bm{e}_{m}$ represents the map registration factor. 
Its derivation is in Sec.~\ref{sec:map_registration}.
The term $\bm{e}_{p}$ represents the marginalization factor.
This marginalization factor encodes information related to old quantities that fall outside the current sliding window.
To compute this factor, we use the strategy proposed by Leutenegger et al.~\cite{forster2016manifold}.
This strategy distinguishes between variables to marginalize (included in the derivation of the term $\bm{e}_{p}$) and variables to drop.
The variables to drop are the ones that are not connected to keyframes, for example, 3D points that are visible from the frame that falls outside the sliding window when this frame is not a keyframe.
Dropping, instead of marginalizing, such variables allows to keep the sparsity of the Jacobian of Eq.~\ref{Eq:VIO_Formulation}.

\subsection{Localization to 3D Digital Twins}\label{sec:map_registration}
\subsubsection{Residual}
At the initialization of our system, a GPS measurement prior is used to access the part of the city digital twin around the current pose.
This also allows to represent the city map in the reference frame $W$.
During deployment, we keep track of the camera pose in both the local frame $L$ and global frame $W$.
This allows to run the map registration algorithm between the sparse local VSLAM point cloud and the global semi-dense city digital twin.
We use the ICP point-to-plane algorithm to solve the registration problem.
Our ICP algorithm minimizes the function:
\begin{equation} \label{eq:min_ICP_function_p2plane}
\min_{\delta \bm{R}, \delta \boldsymbol{p}} \mathcal{J} = \sum_{j\in\mathcal{J}_k} \left( \left[ \delta \bm{R} \cdot \bm{a}_{j} + \delta \bm{p} - \bm{b}_{j} \right] \cdot \bm{n}_j \right)^2
\end{equation}
where $\delta \bm{R}, \delta \bm{p}$ are the rotation and position that align the source point cloud (the local VSLAM point cloud) to the target point cloud (the city digital twin point cloud).
The point $\bm{a}_{j}$ is the point in the source point cloud that is associated to the point $\bm{b}_{j}$ in the target point cloud. 
The quantity $\bm{n}_j$ is the normal vector at the point $\bm{b}_{j}$.

We define the solution of Eq.~\ref{eq:min_ICP_function_p2plane} in the form of a homogeneous transformation matrix a $\delta \bm{T} = [\delta \bm{R} | \delta \bm{p}]$. 
Taken a point in the target point cloud $\bm{m}_{j}^{w}$ and its position derivated from the current pose estimate $\hat{\bm{m}}_{j}^{w}$ (namely, $\bm{m}_{j}^{c} = \hat{\bm{m}}_{j}^{c}$), we derive the global measurement as:
\begin{align}    
    \delta \bm{T} \cdot \hat{\bm{m}}_{j}^{w} &= \bm{m}_{j}^{w} \\
    \delta \bm{T} \cdot \hat{\bm{T}}_{WL} \cdot \hat{\bm{T}}_{{LB}} \cdot \bm{T}_{{BC}} \cdot \bm{m}_{j}^{c} &= \hat{\bm{T}}_{{WL}} \cdot \bm{T}_{{LB}} \cdot \bm{m}_{j}^{w} \\
    \delta \bm{T} \cdot \hat{\bm{T}}_{{WL}} \cdot \hat{\bm{T}}_{{LB}} &= \hat{\bm{T}}_{{WL}} \cdot \bm{T}_{{LB}} \\
    \hat{\bm{T}}_{{LW}} \cdot \delta \bm{T} \cdot \hat{\bm{T}}_{{WL}} \cdot \hat{\bm{T}}_{{LB}} &= \bm{T}_{{LB}},
\end{align}
where $ \bm{T}_{{LB}} = [\bm{R}_{{LB}} | \bm{p}_{{LB}}]$.

The map registration residual term that is added to the optimization problem in Eq.~\ref{Eq:VIO_Formulation} is defined as:
\begin{equation}
\bm{e}_{m}^{k} = \begin{bmatrix}
\hat{\bm{p}}^{{k}}_{{LB}} - \bm{p}^{{k}}_{{LB}} \\
\text{Log}((\hat{\bm{R}}^{{k}}_{{LB}})^t \cdot \hat{\bm{R}}^{{k}}_{{LB}} ),
\end{bmatrix}
\end{equation}
where Log($\cdot$) is the logarithm map that maps an element of the Lie group, \textit{SO}(3), to an element of the Lie algebra.

\subsubsection{Adaptive Weighting}\label{sec:adaptive_weighting_strategy}
The solution of Eq.~\ref{eq:min_ICP_function_p2plane} is not guaranteed to be unique.
The presence of many collinear or coplanar points can lead to unobservable directions.
This situation can be encountered in urban environments, for example when the camera moves in a straight line facing a building facade.
To tackle this problem, we propose an adaptive weighting strategy that assigns a low weight to the detected unobservable direction while still allowing high weights for well-observed directions.
Our solution, inspired by~\cite{Barczyk_ICP}, assigns the weight to the map registration residual, $\text{W}_m^{k}$, in Eq.~\ref{Eq:VIO_Formulation} based on the covariance matrix of the cost function in ~\ref{eq:min_ICP_function_p2plane}.
The cost function in~\ref{eq:min_ICP_function_p2plane} can be linearized as:
\begin{equation}
\min_{\bm{x}} \sum_{j\in\mathcal{J}_k} \left|\left| \bm{y_j} - \bm{A_j} \bm{x} \right|\right|^2,
\end{equation}
where:
\begin{equation}
\bm{y}_j = \bm{n}_j^T (\bm{a}_j - \bm{b}_j), \quad \bm{A}_j = \begin{bmatrix} -(\bm{a}_j \times \bm{n}_j)^T \ -\bm{n}_j^T \end{bmatrix}.
\end{equation}
The covariance is then equivalent to the inverse of the Hessian, \( \bm{H} = \sum_{j} (\bm{A_j}^T \bm{A_j}) \). 
The weight of the map registration residual is defined as:
\begin{equation}
\label{eq:weightICP}
\begin{aligned}
\bm{W}_m^k &= \frac{\beta }{\sigma^2} \cdot \exp\left(-\frac{\gamma^2}{2}\right) \cdot \bm{H}, \\
\end{aligned}
\end{equation}
where: $\gamma$ is the root-mean-square-error of the inliers, $\sigma^2 = \sum_{k} (\bm{H}_{kk})$ is the sum of the diagonal elements of $\bm{H}$, and $\beta$ is a scaling factor that makes $\bm{W}_m^k$ of the same order of magnitude as the weights of the other residuals in Eq.~\ref{Eq:VIO_Formulation}.
We found $\beta$ empirically and kept it constant in all our experiments.

\subsection{Initial Frame alignment}\label{sec:FrameAlignement}

The relative transformation (position, and heading, i.e., the rotation around the gravity-aligned z-axis), $\bm{T}_{WL} = [\bm{R}^{z}_{WL} | \bm{p}_{WL}]$ is estimated as new position estimates from the VSLAM, $\bm{p}^{k}_{LB}$, and global measurements, $\bm{p}^{k}_{WB}$, are available. 
Our frame alignment solution is found by optimizing the cost function: 
\begin{equation} \label{eq:FrameAlignement}
\arg \min_{\bm{R}_{WL}, \bm{t}_{WL}} \sum_{n} \| \bm{p}^{n}_{WB} - \bm{R}^{z}_{WL} \cdot \bm{p}_{LB} - \bm{p}_{WL} \|^2
\end{equation}
We apply the Umeyama method~\cite{Umeyama1991LeastSquaresEO} to find the solution to Eq.~\ref{eq:FrameAlignement}. 
Initially (before map registrations can be successfully found), we solve the problem in Eq.~\ref{eq:FrameAlignement} using the VSLAM pose estimates and the GPS measurements.
Due to the inaccuracy of GPS measurements in urban environments, this initial alignment can be inaccurate.
We found empirically on real data that the heading error resulting from the frame alignment based on GPS measurements cannot be neglected.
Once the map registration can be successfully performed, we use the position obtained from the map registration, c.f. Eq.~\ref{eq:min_ICP_function_p2plane}, instead of the GPS data.
The frame alignment Eq.~\ref{eq:FrameAlignement} is run until a convergence criterion is met.
We adopt the convergence criteria proposed by Boche et al.~\cite{boche2022visualinertial}, which checks that the heading covariance is below a threshold.

%



%
%
%



\subsection{Implementation Details}

We integrated the proposed geometry-based map registration strategy in SVO~\footnote{\url{https://github.com/uzh-rpg/rpg_svo_pro_open}}.
The sliding-window optimization is based on the approach presented in~\cite{OKVIS_2013}. 
We limited the number of keyframes in the sliding-window to 5.
The visual frontend is based on SVO~\cite{Forster17troSVO}.
\rebuttal{We run the ICP registration on a local portion, of constant size, of the global map. 
Consequently, the runtime and memory requirements of the map registration process do not depend on the dimensions of the global map.
In all our experiments, the local portion of the global map has horizontal dimensions of 150 x 150 m around the current camera position.
We take all the available points along the vertical dimension.}
We name our system SVO-Digital Twin.
All the experiments ran on a laptop equipped with a 2.60GHz Intel Core i7 CPU.
Our system runs in real time, namely, it can complete the processing of one image before the next one is available (the maximum camera frame rate in all the datasets used in the experiments is 30 Hz).

%% file: sections/experiments.tex
\section{Experiments}\label{sec:Experiements}
\begin{figure*}
    \centering
    \includegraphics[width=1.0\linewidth]{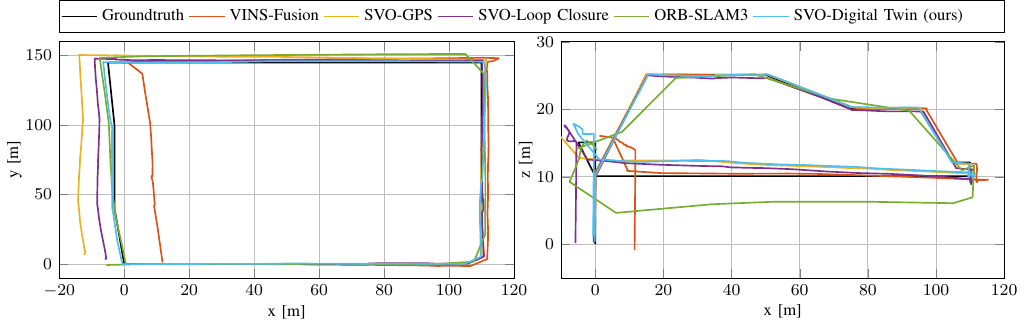}
    \caption{Simulated data. Trajectory estimated by SVO-Digital Twin and the baselines together with the ground truth trajectory. Left: Top-down view, right: Side view.}
    \label{fig:simulation}
\end{figure*}
To evaluate the effectiveness of our method, we conducted experiments in both simulated and real-world environments. 
We compare our method against four baselines. 
To compare our method against VIO systems that integrate GPS data, we select VINS-Fusion~\cite{qin2019general} and SVO-GPS~\cite{cioffi2020tightly} as baselines.
VINS-Fusion loosely couples a VIO~\cite{VinsMono} with GPS measurements.
Instead, SVO-GPS tightly couples the GPS measurements in the VIO sliding-window optimization.
To compare our method against SLAM systems that close loops based on visual feature matching, we select SVO-Loop Closure~\cite{Forster17troSVO} and ORB-SLAM3~\cite{campos2021orb}.
SVO-Loop Closure uses the same VIO pipeline as our system.
ORB-SLAM3 is a Visual SLAM system that performs loop closure by matching ORB features.

In all our experiments, we use the metrics:
\begin{itemize}
    \item Positional absolute trajectory error ($\text{ATE}_{\text{P}}$) [m] : $\sqrt{\frac{1}{N} \sum_{k=0}^{N-1} \norm{ \bm{p}^k_{WB} - \hat{\bm{p}}^k_{WB}}^2 }$. 
    \item Rotational absolute trajectory error ($\text{ATE}_{\text{R}}$) [deg] : $\sqrt{\frac{1}{N} \sum_{k=0}^{N-1} \norm{ \text{Log}((\bm{R}^{k}_{WB})^t \cdot \hat{\bm{R}}^k_{WB} ) }^2 }$.
\end{itemize}
We use the trajectory evaluation toolbox proposed by Zhang et al.~\cite{Zhang18iros} to compute these evaluation metrics.
We use the transformation $\bm{T}_{WL}$, c.f. Eq~\ref{eq:FrameAlignement}, to align the pose estimates to the world frame and then compute the evaluation metrics.
This transformation is computed offline, as proposed in~\cite{Zhang18iros}, for the systems that do not use global measurements.

\subsection{Simulation results}
\subsubsection{Setup} 
We use the trajectory simulator in~\cite{Foehn_2022} to simulate trajectories and IMU data, and, the Flightmare simulator~\cite{song2020flightmare} to render images from a virtual city model.

Differently from the related works~\cite{cioffi2020tightly, qin2019general, boche2022visualinertial}, where GPS measurements are generated by sampling a zero-mean Gaussian distribution with a pre-selected standard deviation, we train a data-driven measurement generator on real GPS data.
We collected training data by randomly sampling poses in our simulated city.
We assume that the simulated city is located at the same coordinates as Zurich, Switzerland.
We use ray tracing~\cite{ollander} for each data sample to compute the received GPS signals.
In this way, we create two datasets, one containing the number of satellites seen as a function of the height, and the other containing the multipath error of a single satellite as a function of the height.
This dataset generation strategy is inspired by~\cite{ollander}.
We trained a Gaussian Process (GP) regressor to predict the number of visible satellites and a Gaussian Mixture Model (GMM) to predict the multipath error as a function of the height.
To generate data for the VIO/SLAM algorithms, we simulate a drone flying a trajectory long 500 meters.
During the flight, the GP regressor is used to predict the number of visible satellites, and the GMM is used to generate a multipath error for each satellite.
The GPS measurement is then generated by performing trilateration using the measurements from all the visible satellites.
GPS measurements are generated at 5 Hz.
Our virtual city is shown in Fig.~\ref{fig:simulated_city}. 
The simulated GPS data is shown in Fig.~\ref{fig:simulated_gps}. 
\input{figures/simulated_city}
\input{figures/figureSimulation/simulated_gps}
\input{tables/simulation_results}
\input{tables/ablation_adaptive_weight}
\input{tables/runtime}
\begin{figure*}
    \centering
    \includegraphics[width=\linewidth]{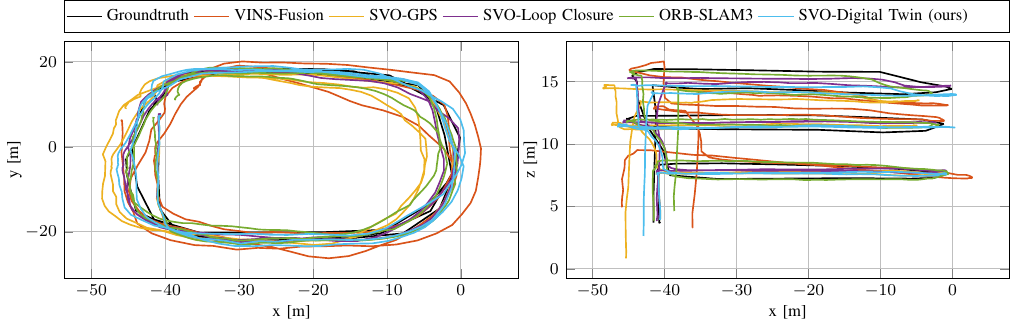}
    \caption{Real-world data. Trajectory estimated by the SVO-Digital Twin and the baselines together with the ground truth trajectory. Left: Top-down view, right: Side view.}
    \label{fig:realworld}
\end{figure*}
\input{tables/realworld_results}
\subsubsection{Results}
The absolute trajectory errors achieved by our system, SVO-Digital Twin, and the baselines are reported in Table~\ref{tab:simulation_results}.
The estimated trajectories are plotted, together with the ground truth, in Fig~\ref{fig:simulation}.
SVO-Digital Twin achieves the best absolute trajectory error in position and rotation by a large margin.
Specifically, the $\text{ATE}_{\text{p}}$ and $\text{ATE}_{\text{R}}$ improve by 31\% and 27\%, respectively, with respect to the best-performing baselines.
%

We include in Table~\ref{tab:ablation_adaptive_weighting} an ablation study to validate our adaptive weighting strategy.
The improved ATE errors validate the benefits of the proposed adaptive weighting strategy.

\rebuttal{We include the runtime of the main system components in Table ~\ref{tab:runtime}. 
We fix a time length for the initial alignment step.
After this time has passed, we check if the convergence criterion (see, Sec.~\ref{sec:FrameAlignement}) is met. 
If yes, the alignment is deemed successful.
If not, we accumulate more GPS measurements until this criterion is met.
The runtime of pose estimation is the time elapsed from when an image is captured until the camera pose of such an image is estimated.
}



\subsection{Real world experiments}
\subsubsection{Setup}
\input{figures/real_city}
\input{figures/view_comparison}
We flew a Flyability Elios 3~\footnote{\url{https://www.flyability.com/elios-3}}, equipped with a GPS antenna, a monocular fisheye camera, and an IMU in a neighborhood featuring tall buildings in Zurich, Switzerland.
The flight path lasts circa 5 minutes and covers over 500 meters.
The drone altitude varies between 0 and 17 meters.
The drone flies three loops around a building at different altitudes, which increase by approximately 4 meters per loop.
The drone maintained a 45-degree angle towards the building during most of the flight, ensuring visibility of both the structure and the surrounding environment.
The ground truth was generated using a structure-from-motion framework, COLMAP~\cite{schoenberger2016sfm}.
We use the city digital twin from Google Photorealistic 3D Tiles~\footnote{\url{https://3d-tiles.web.app}}, c.f. Fig.~\ref{fig:real_city}.
The camera trajectory obtained from COLMAP is aligned to the world frame by using an ICP algorithm that takes as inputs the point cloud generated by COLMAP and the point cloud of the city map.
We show 2 images taken at the same location but in two different laps in Fig.~\ref{fig:view_comparison}.
The different camera point of view makes visual loop closure difficult, indeed, neither SVO Loop Closure nor ORB-SLAM3 can close a loop in this case.
Contrary, our map registration is performed successfully in this situation.


\subsubsection{Results}
The absolute trajectory errors achieved by our system, SVO-Digital Twin, and the baselines are reported in Table~\ref{tab:realworld_results}.
The estimated trajectories are plotted, together with the ground truth, in Fig~\ref{fig:realworld}.
SVO-Digital Twin achieves the best absolute trajectory error in position and rotation by a large margin.
Specifically, the $\text{ATE}_{\text{p}}$ and $\text{ATE}_{\text{R}}$ improve by 32\% and 28\%, respectively, with respect to the best-performing baselines.
Since SVO-GPS tightly couples the GPS with visual and inertial data, its performance is degraded due to the high uncertain GPS measurements.

Furthermore, we evaluated the frame alignment strategy proposed in Sec~\ref{sec:FrameAlignement} against aligning the VSLAM pose estimates with the world-frame referenced ground truth poses.
Our alignment strategy achieves a translation error of 0.50 [m] and a heading error of 0.79 [deg].
Otherwise, using only GPS measurement in our alignment strategy results in a translation error of 8.11 [m] and a heading error of 6.77 [deg].

%% file: figures/simulated_city.tex
\begin{figure}
    \centering
    \includegraphics[width=\linewidth]{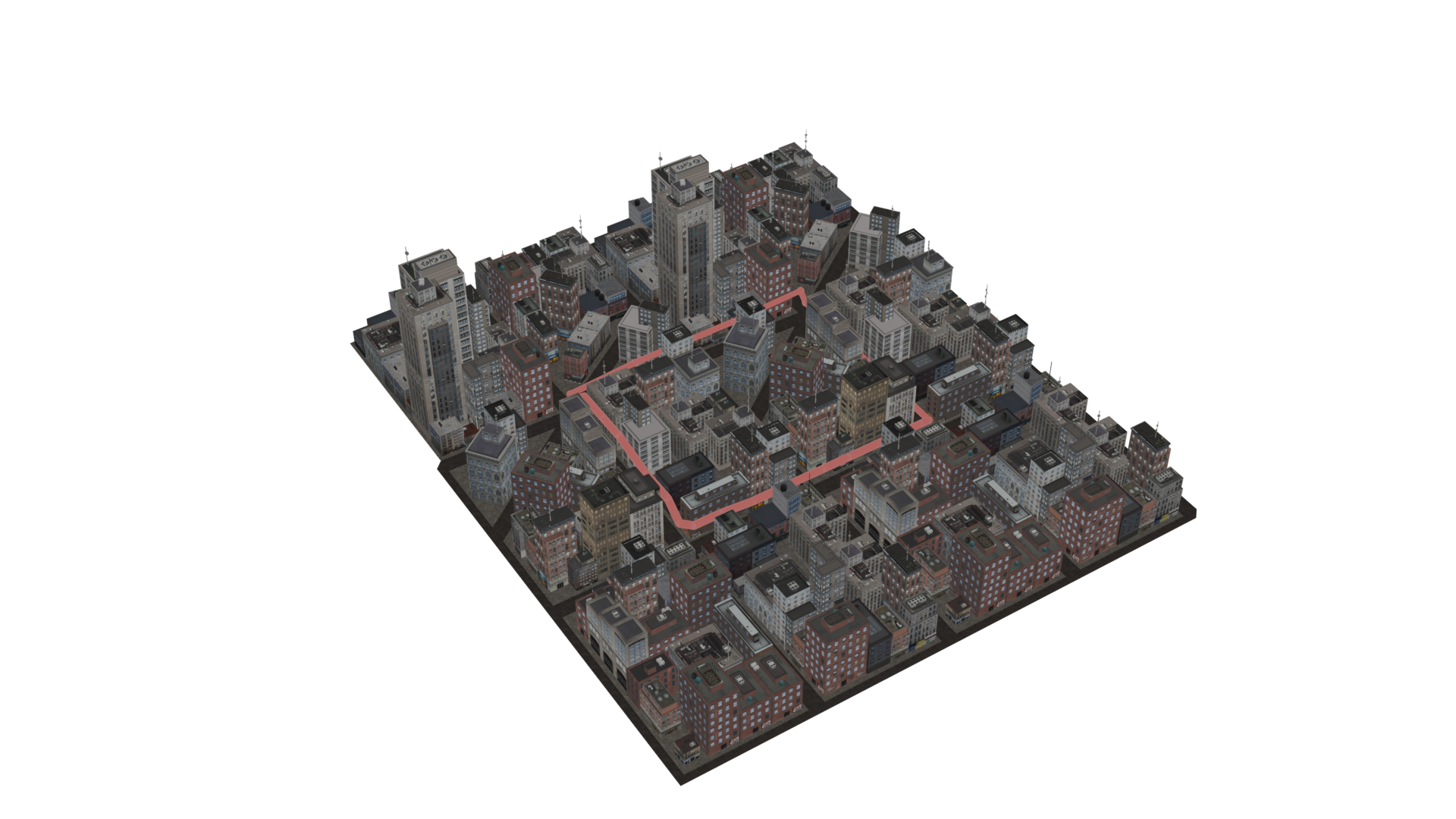}
    \caption{View of the simulated city. The simulated trajectory is depicted in red.}
    \label{fig:simulated_city}
\end{figure}

%% file: figures/figureSimulation/simulated_gps.tex
\begin{figure}
    \centering
    \includegraphics[width=\linewidth]{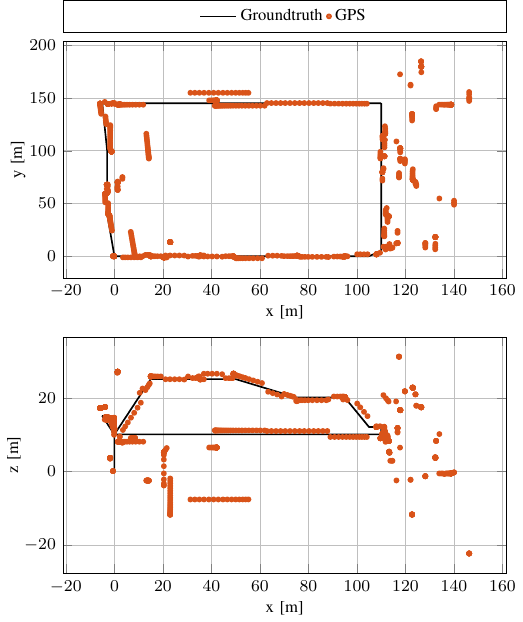}
    \caption{Visualization of the simulated GPS measurements.}
    \label{fig:simulated_gps}
\end{figure}

%% file: tables/simulation_results.tex
\begin{table}
\caption{Evaluation of the trajectory estimates in the simulated dataset.
In bold are the best values.}
\vspace{3pt}
\label{tab:simulation_results}
\setlength{\tabcolsep}{4pt}
\begin{tabularx}{1.0\linewidth}{P{2.0cm}|C|C|C|C|C}
\toprule 
\multirow[c]{2}{=}[-6pt]{\centering Evaluation \newline Metric} & \multicolumn{4}{c}{VIO / SLAM Algorithm}  \\[4pt]
& Vins-Fusion & SVO-GPS & SVO-Loop Closure & ORB-SLAM3 & SVO-Digital Twin (Ours) \\
 \midrule
 $\text{ATE}_{\text{P}}$ [m] & 7.87 & 7.84 & 4.02 & 4.32 & \textbf{2.78} \\[4pt]
 $\text{ATE}_{\text{R}}$ [deg] & 1.30 & 1.22 & 1.49 & 1.49 & \textbf{0.89} \\
\bottomrule
\end{tabularx}
\end{table}

%% file: tables/ablation_adaptive_weight.tex
\begin{table}
\caption{Evaluation of the benefit of the adaptive weighting strategy presented in Sec.~\ref{sec:adaptive_weighting_strategy} }
\vspace{3pt}
\label{tab:ablation_adaptive_weighting}
\setlength{\tabcolsep}{4pt}
\begin{tabularx}{1.0\linewidth}{P{2.0cm}|C|C}
\toprule 
\multirow[c]{2}{=}[-6pt]{\centering Evaluation \newline Metric} & \multicolumn{2}{c}{SVO-Digital Twin}  \\[4pt]
& w/o Adaptive Weighting & w/ Adaptive Weighting (Ours) \\
 \midrule
 $\text{ATE}_{\text{P}}$ [m] & 4.54 & \textbf{2.78} \\[4pt]
 $\text{ATE}_{\text{R}}$ [deg] & 2.41 & \textbf{0.89} \\
\bottomrule
\end{tabularx}
\end{table}

%% file: tables/runtime.tex
\begin{table}
\caption{\rebuttal{Runtime of the main system components.}}
\vspace{3pt}
\label{tab:runtime}
\setlength{\tabcolsep}{4pt}
\begin{tabularx}{1.0\linewidth}{P{2.0cm}|C}
\toprule 
{\centering System component} & {\centering Average Runtime [s]} \\
 \midrule
 Initial alignment & 40.0 \\[2pt]
 ICP-registration & 0.44 \\[2pt]
 Pose estimation & 0.03 \\
\bottomrule
\end{tabularx}
\end{table}

%% file: tables/realworld_results.tex
\begin{table}
\caption{Evaluation of the trajectory estimates in the real-world dataset.
In bold are the best values.}
\vspace{3pt}
\label{tab:realworld_results}
\setlength{\tabcolsep}{4pt}
\begin{tabularx}{1.0\linewidth}{P{2.0cm}|C|C|C|C|C}
\toprule 
\multirow[c]{2}{=}[-6pt]{\centering Evaluation \newline Metric} & \multicolumn{4}{c}{VIO / SLAM Algorithm}  \\[4pt]
& Vins-Fusion & SVO-GPS & SVO-Loop Closure & ORB-SLAM3 & SVO-Digital Twin (Ours) \\
 \midrule
 $\text{ATE}_{\text{P}}$ [m] & 1.61 & 3.05 & 1.60 & 2.36 & \textbf{1.09} \\[4pt]
 $\text{ATE}_{\text{R}}$ [deg] & 11.30 & 5.99 & 4.70 & 7.04 & \textbf{3.40} \\
\bottomrule
\end{tabularx}
\end{table}

%% file: figures/real_city.tex
\begin{figure}
    \centering
    \includegraphics[width=\linewidth]{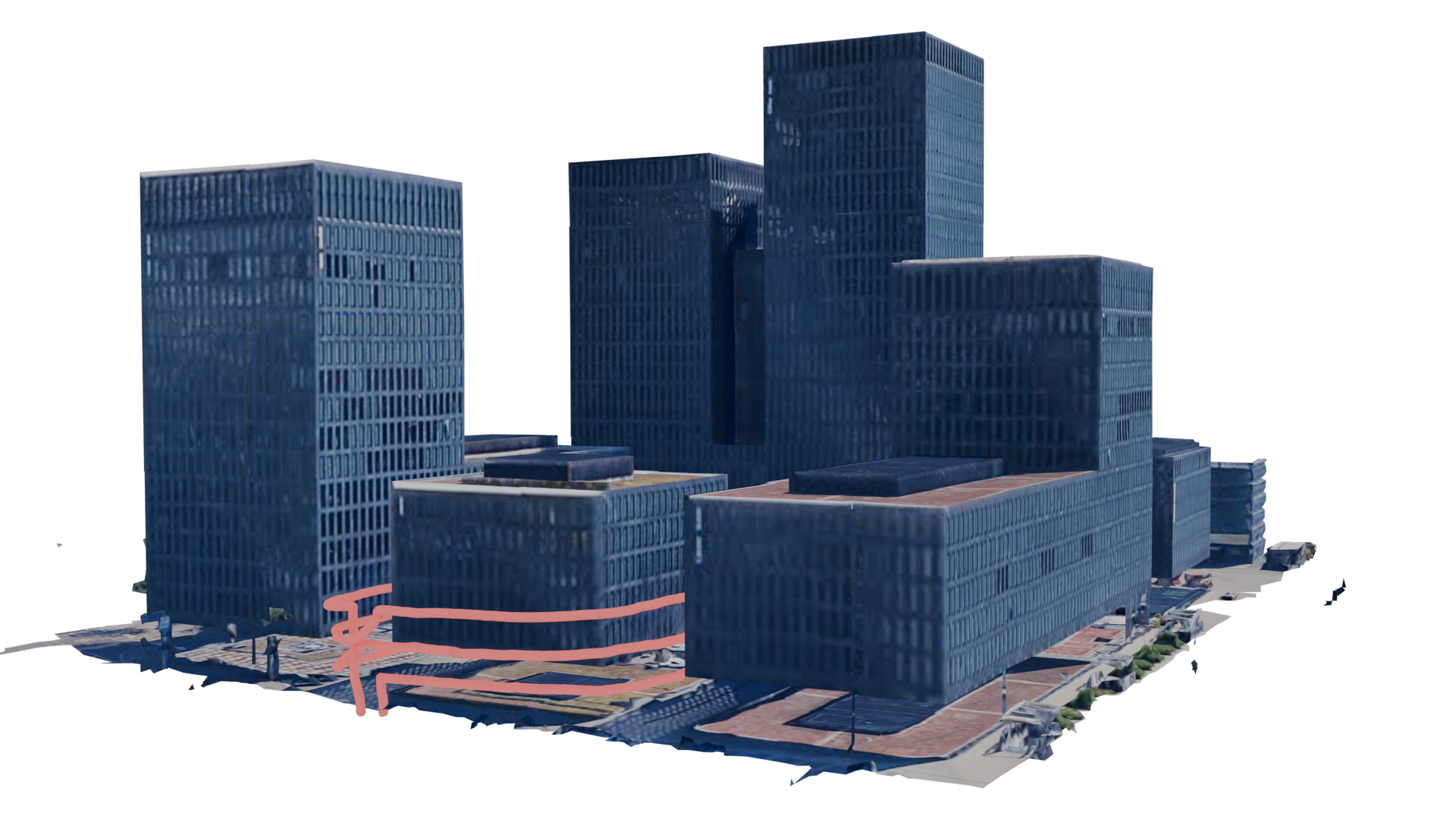}
    \caption{Rendered map from Google Photorealistic 3D Tiles of the neighborhood in Zurich where we collected the real-world data. The trajectory flown by our drone is depicted in red.}
    \label{fig:real_city}
\end{figure}

%% file: figures/view_comparison.tex
\begin{figure}
    \centering
    \includegraphics[width=\linewidth]{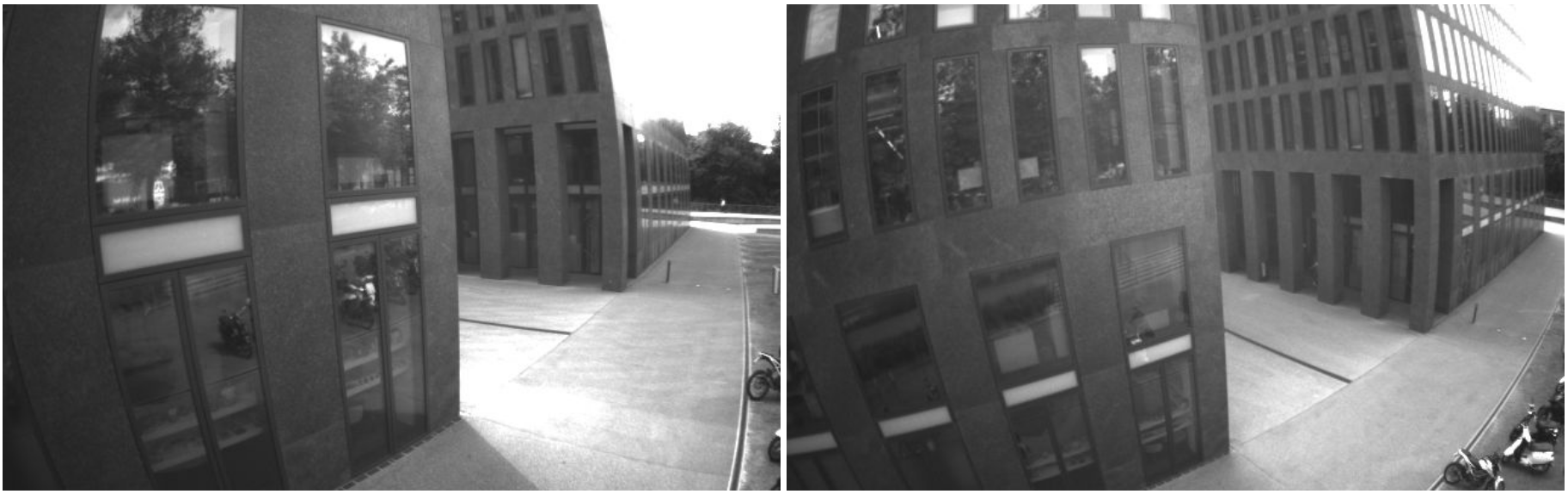}
    \caption{Images of the same part of the building taken in two consecutive laps. Neither SVO-Loop Closure nor ORB-SLAM3 can close a loop based on visual descriptor matching. Contrary, our map registration is performed successfully in this situation.}
    \label{fig:view_comparison}
\end{figure}

%% file: sections/discussion.tex
\section{Discussion}\label{sec:discussion}

\rebuttal{In this section, we discuss the failure cases of the global map registration and the generalizability of our approach to any \textit{Digital Twin}.

The registration between the local VSLAM point cloud and the global map can be incorrect for two main reasons.
The first reason is 3D aliasing. 
Our adaptive weighting strategy can detect this failure case and provide a solution, c.f. Tab.~\ref{tab:ablation_adaptive_weighting}.
The second reason is inaccuracies in either the global map or the local 3D point cloud that lead to incorrect matches in the ICP-based registration.
Inaccuracies in the global map are mainly the consequence of changes in the environment.
This case is detected by the outlier rejection strategy that is part of our ICP algorithm, and the registration is deemed unsuccessful.
Inaccuracies in the local 3D point cloud are due to large drift that could accumulate in the VSLAM system after a long period of unsuccessful global map registrations. 
In this case, the solution is to re-align the local VSLAM reference frame to the global frame using GPS or any other global measurement. 
A possible implementation of such a solution can be based on the approach proposed in~\cite{boche2022visualinertial}, where the VIO local frame is re-aligned to the global frame after a long time of GPS outage.

Our method is not restricted to the availability of Google 3D tiles maps.
Indeed, it can work with any \textit{Digital Twin} that represents the geometry of the environment via 3D points, meshes, etc.
In this case, the camera's initial position in the Digital Twin needs to be known and it can be set manually. 
Google 3D tiles are one possible representation of a city's Digital Twin.
Another example of the city's Digital Twin that can be used in our system is the open-source maps provided by OpenStreetMap~\footnote{\url{https://www.openstreetmap.org}}.
OpenStreetMap maps provide 3D information of the buildings but they do not have texture information included in most places.
A method relying on image features would not be able to use OpenStreetMap maps.
}

%% file: sections/conclusion.tex
\section{Conclusion}\label{sec:conclusion}

In conclusion, this paper presents a novel approach to reduce the drift of VIO/VSLAM systems by leveraging geometric information from digital twins. 
By introducing a method that localizes the VIO/VSLAM-generated point cloud to a city digital twin via point-to-plane matching, we address the limitations of methods that rely on GPS or visual localization techniques. 
Our experiments, both in a high-fidelity GPS simulator and real-world drone flight, demonstrate the superiority of this method over existing systems, particularly in reducing drift and handling viewpoint variations. 
%



